\documentclass[letterpaper, 10 pt, conference]{ieeeconf}  

\IEEEoverridecommandlockouts                              

\usepackage{moreverb,url,multicol,amsmath,subfigure,bm,algorithm,algpseudocode,multirow,empheq,graphicx,amssymb,xcolor,siunitx, soul,booktabs,tabularx,makecell, url, hyperref}

\title{\LARGE \bf
CCDP: Composition of Conditional Diffusion Policies\\with Guided Sampling}

\author{Amirreza Razmjoo$^{1,2,3}$, Sylvain Calinon$^{2,3}$, Michael Gienger$^{1}$, and Fan Zhang$^{1}$ \\
\thanks{$^{1}$Honda Research Institute Europe GmbH, Germany 
        {\tt\small \{michael.gienger, Fan.zhang\}@honda-ri.de}}
\thanks{$^{2}$Idiap Research Institute, Switzerland 
 {\tt\small\{amirreza.razmjoo, Sylvain.calinon\}@idiap.ch}}%
\thanks{$^{3}$École Polytechnique Fédérale de Lausanne (EPFL), Switzerland      
}}

\begin{document}

\maketitle
\thispagestyle{empty}
\pagestyle{empty}

\begin{abstract}
Imitation Learning offers a promising approach to learn directly from data without requiring explicit models, simulations, or detailed task definitions. During inference, actions are sampled from the learned distribution and executed on the robot. However, sampled actions may fail for various reasons, and simply repeating the sampling step until a successful action is obtained can be inefficient. In this work, we propose an enhanced sampling strategy that refines the sampling distribution to avoid previously unsuccessful actions. We demonstrate that by solely utilizing data from successful demonstrations, our method can infer recovery actions without the need for additional exploratory behavior or a high-level controller. Furthermore, we leverage the concept of diffusion model decomposition to break down the primary problem—which may require long-horizon history to manage failures—into multiple smaller, more manageable sub-problems in learning, data collection, and inference, thereby enabling the system to adapt to variable failure counts. Our approach yields a low-level controller that dynamically adjusts its sampling space to improve efficiency when prior samples fall short. We validate our method across several tasks, including door opening with unknown directions, object manipulation, and button-searching scenarios, demonstrating that our approach outperforms traditional baselines. Supplementary materials for this paper are available on our website: {\color{red} \href{https://hri-eu.github.io/ccdp/}{https://hri-eu.github.io/ccdp/}.}
\end{abstract}

\section{INTRODUCTION}
Recent advances in imitation learning—exemplified by methods such as Implicit Behavior Cloning \cite{florence2022implicit} and diffusion/flow-matching policies \cite{chi2023diffusion,zhang2024affordance}—have demonstrated remarkable capabilities in learning and replicating complex data distributions from demonstrations. These approaches effectively capture the underlying distribution of expert behaviors, enabling robots to generate actions by sampling from this learned distribution at inference time. Despite their success, a fundamental challenge remains: What should the robot do when a sampled action fails? Just as humans often encounter small failures (like fumbling for a bedside lamp switch in the dark or pushing a door in the wrong direction), robots operating in partially constrained environments may also face situations where certain actions become infeasible or ineffective. In scenarios where the robot only has access to the successful expert demonstrations (which is usually the case), the robot must be designed to not simply repeat failed attempts but instead to actively explore alternative actions seen in the demonstrations, including infrequent ones.


Existing approaches to failure recovery often assume access to supplementary resources—such as simulated environments \cite{lee2019robust}, advanced reasoning or foundational models \cite{duan2024aha}, and expert guidance \cite{niekum2013incremental}—which may not be readily available in practical settings. A considerable body of research has tackled this problem using a two-level planning approach, where a high-level planner, such as foundational models for reasoning \cite{duan2024aha,huang2022inner,liu2023reflect} or two-level policies trained through reinforcement learning (RL) or behavior cloning (BC) \cite{lee2019robust,triantafyllidis2023roman}, selects action primitives and task parameters based on past experiences. While promising, this approach has notable drawbacks. First, multi-layer planners can lead to suboptimal results due to the added complexity and separation between planning levels. Second, when the number of options increases, the system can suffer from \emph{combinatorial explosion}, making decision-making inefficient and computationally expensive. For these reasons, we adopt an alternative approach to handling failure recovery directly within the low-level controller, which is the primary focus of this paper. Moreover, in some failure cases, like the bedside lamp example, there is often little room for reasoning. If previous attempts to locate the switch fail, the only useful information gained is that the switch is not in the already-checked positions. Inspired by this observation, this paper investigates how such failure-driven behavior guidance can be implemented directly in robot policies.

\begin{figure}[t!]
   \centering
    \subfigure{%
        \includegraphics[width=0.45\textwidth]{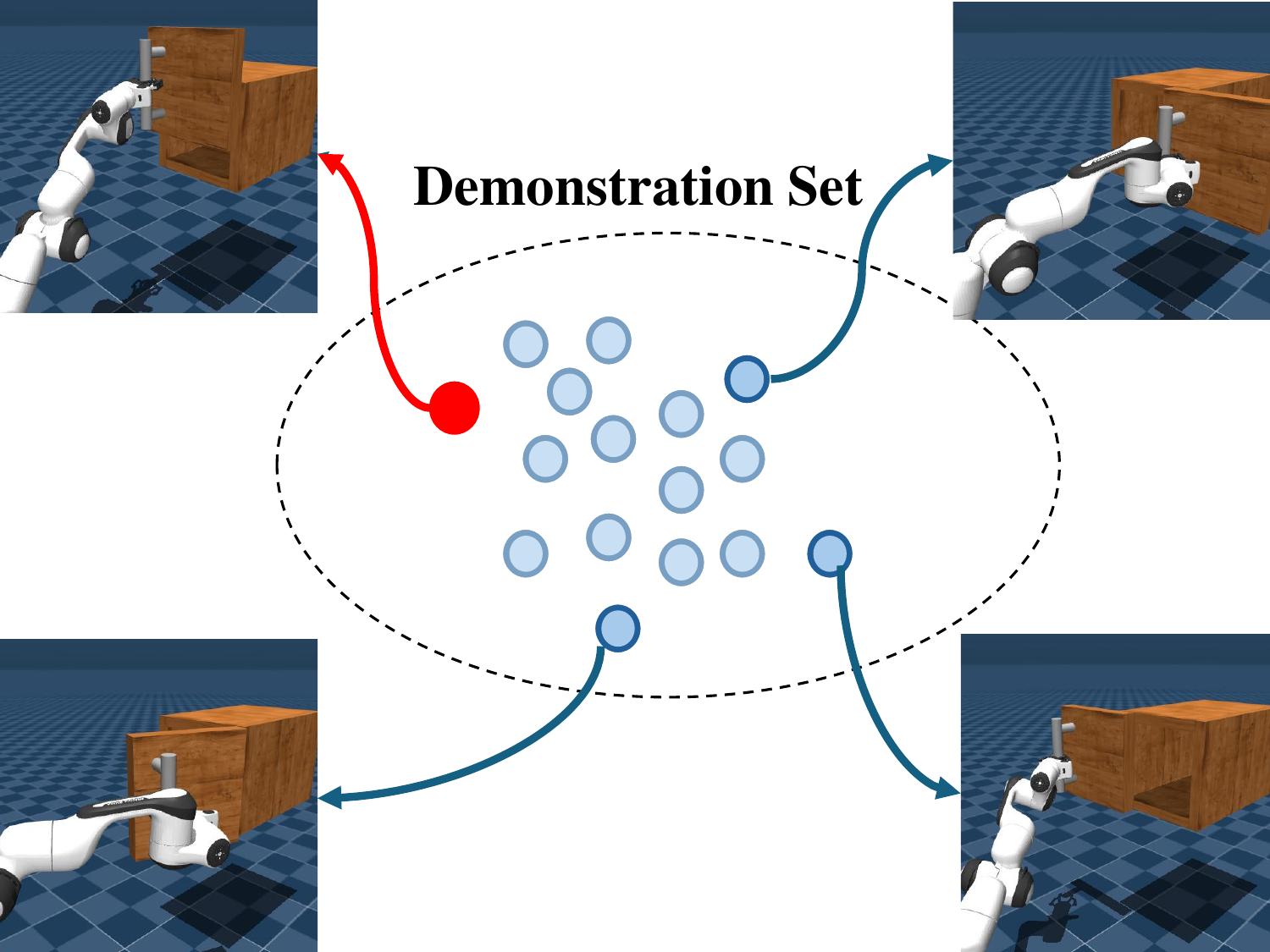}
        }
     \caption{A diverse demonstration set, featuring multiple task variations, is provided to the robot. In the event of a failure, the robot switches to alternative variations rather than repeatedly sampling the same actions.}\vspace{-15pt}
     \label{fig:exp_proces}
\end{figure}

On the low-level controller side, two main approaches are generally considered. The first is to learn a policy that is robust to failures, similar to methods in robust reinforcement learning \cite{chebotar2019closing,tobin2017domain,Xue2024DA} or model predictive control (MPC) \cite{bemporad2007robust}. While this approach can handle some types of failures, like those caused by environmental parameter shifts (e.g., changes in friction), it struggles with a wider range of failures where a single robust policy may not exist. In the button-seeking example, no single action reliably locates the switches, which may be positioned arbitrarily within the search area. Moreover, robust policy methods often require prior task knowledge, such as access to a model or simulator for training, which can be a significant limitation.

Due to these challenges, we focus on the second approach: training a policy that considers a history of observations and actions, updating its decisions based on past outcomes \cite{Lee2020TeacherStudent,Yu2017PreparingFT}. However, this approach also comes with its own difficulties. First, for the policy to learn failure recovery, the system must either encounter failures during demonstrations—making data collection more complex—or explore possible failure scenarios autonomously, which requires either modeling the task environment or extensive, potentially dangerous real-robot training. In this work, we propose a simpler assumption that notably improves system success rates without these heavy requirements. Specifically, we assume that although the executed action failed, a viable recovery action exists in the demonstration set. To leverage this assumption, we perform an offline analysis of the demonstration set to identify a collection of actions that are not similar to each other and train a model that guides the controller to sample an action that is sufficiently \emph{far} from the failed one. Although straightforward, this approach substantially enhances performance.

The second challenge involves capturing failure events over extended histories. Effective failure recovery requires observing multiple successive failures, which necessitates maintaining a long memory of past actions and observations. We address this challenge by leveraging the concept of ''composition of different models'' \cite{liu2022compositional}, also known as the product of experts \cite{hinton1999poe}. In our approach, we independently learn recovery strategies for individual failure events. During inference, these recovery distributions are combined to form a new distribution that effectively compensates for all observed failures. It is important to note that this method assumes the underlying cause of failure remains static over time, making it particularly effective for failures with unchanging roots. Ultimately, the proposed method enables efficient and adaptive guidance of the robot’s action distribution.

In this paper, we introduce the Composition of Conditional Diffusion Policies (CCDP), which builds on a standard diffusion policy framework \cite{chi2023diffusion} but conditions the policy on key failure cases identified by a failure detection system. We assume that the robot is provided with examples of successfully performing a task, but during execution, some actions may fail. When this happens, CCDP guides the robot to select another action that is reasonably far from the previously failed attempts. Unlike traditional conditional models that attract the policy toward desirable behaviors, CCDP uses failure cases as negative guidance, steering the policy away from regions associated with unsuccessful outcomes. This formulation is analogous to the distinction between goal-reaching and obstacle avoidance—the latter often being more challenging in robotic applications. To manage the complexity of failure recovery, such as handling long action histories, we further decompose the task using a composition of diffusion models, effectively breaking the overall task into multiple, more manageable sub-problems. Our approach eliminates the need to classify data into multiple categories, reducing the burden of data labeling and enabling the system to capture additional variations during task reproduction.

In summary, our contributions in this paper are threefold:
\begin{itemize} 
\item We propose a decomposed diffusion policy that enhances the modularity and controllability of the method introduced in \cite{chi2023diffusion}, and we analyze the impact of these modules. 
\item We introduce a recovery strategy that integrates with the base module, enabling the system to recover from unsuccessful attempts without requiring data labeling or environment modeling. 
\item We validate the proposed method through comprehensive comparisons with state-of-the-art baselines. \end{itemize}

\section{Related Works}
\label{sec:rel}
\textbf{Imitation Learning:} Imitation learning is a powerful technique for teaching robots to perform tasks by directly imitating expert behavior \cite{Osa2018AnAP}. This approach offers several advantages, such as eliminating the need for explicit system modeling, cost function design, and extensive manual coding. Moreover, this method has been successfully integrated with reinforcement learning \cite{Nair2017OvercomingEI} and optimal control techniques \cite{razmjoo2021optimal}, enhancing their performance in complex tasks. In our work, we focus exclusively on leveraging expert demonstrations, assuming no access to an explicit cost function or physical model (even simulation environment) during the training phase. Experts provide a distribution of possible solutions for a given task, which motivates using generative models to capture this distribution and sample actions during execution. Early approaches, such as Probabilistic Movement Primitives (ProMP) \cite{paraschos2013probabilistic} and Task-Parameterized Gaussian Mixture Models (TP-GMM) \cite{Calinon16JIST}, laid the groundwork for this framework. More recent methods, including Implicit Behavior Cloning \cite{florence2022implicit}, diffusion policies \cite{chi2023diffusion}, and flow-matching policies \cite{zhang2024affordance}, have further advanced the field by modeling demonstration distributions more effectively. Despite these advances, it is not guaranteed that a single sampled action will succeed. A common strategy to address this issue is to repeatedly sample from the same distribution until a successful action is obtained. However, this approach can be inefficient as it does not account for the fact that certain regions of the action space have already been identified as unsuccessful. In contrast, our work seeks to explicitly guide the system away from regions associated with failure, thereby improving both efficiency and overall success rates.

\textbf{Guided Policy Inference:}
Several approaches have been proposed to guide the inference process. One class of methods conditions policy parameters on system characteristics, updating these parameters through the robot’s interactions with the environment \cite{ADETOLA2009320,xue2024robust}. These techniques leverage system identification or adaptive control strategies to adjust the sampling process. However, because they depend on an underlying model, they may not be directly applicable in imitation learning scenarios where such models are unavailable. An alternative strategy involves using high-level decision variables controlled by an external module, such as foundational models \cite{huang2022inner,xie2024deligrasp} or controllers trained via hierarchical reinforcement learning (HRL) \cite{lee2019robust} or hierarchical behavior cloning (HBC) \cite{triantafyllidis2023roman}. Although effective in some contexts, this approach can lead to an explosion in the number of decision variables for the high-level planner—a problem commonly referred to as \emph{combinatorial explosion}—which makes decision-making computationally expensive. A further approach is to employ rejection sampling, a Monte Carlo method where failed samples are discarded and new ones are generated. In our work, rather than rejecting individual samples, we reject entire sampling regions associated with failure. This modification reduces the likelihood of repeatedly sampling from unproductive areas. We achieve this goal through the composition of multiple diffusion models, enabling efficient and adaptive exploration of the action space without relying on a high-level decision maker.

\textbf{Composition of Multiple Models:}
The concept of Products of Experts (PoE), introduced by Hinton \cite{hinton1999poe}, suggests that instead of learning a single complex distribution over many variables, one can decompose the problem into several simpler subproblems (i.e., ''experts''), each focusing on a specific objective and/or variables. This framework has been successfully applied in robotics, for example, to fuse sensory information \cite{pradalier2003expressing}, combine multiple coordinate systems in imitation learning \cite{Calinon16JIST}, and integrate optimality with feasibility in model predictive control \cite{razmjoo2024sampling}. The emergence of energy-based models \cite{lecun2006tutorial} and later denoising models, such as those in \cite{ho2020denoising,lipman2022flow}, has further popularized the use of PoE, often referred to in these contexts as the composition of models. These approaches are particularly adept at handling high-dimensional problems with complex distributions. Recent works \cite{gkanatsios2023energybased,yang2023diffusion} have demonstrated how combining multiple constraint models via a products-of-experts approach can effectively address challenging task and motion planning problems. In this work, we extend this idea to the low-level controller, guiding the inference by composing multiple diffusion models to avoid the regions associated with failure.

\section{Preleminaries}
\label{sec:pre}
\subsection{Diffusion Policy}\label{sec:pre_dp}
Let \(\mathbf{a}_t \in \mathbb{R}^{d_u}\) denote the action at time \(t\), \(\mathbf{x}_t \in \mathbb{R}^{d_s}\) the state, and \(\mathbf{h}_t^H = [\mathbf{a}_{t-H:t-1}^\top, \mathbf{x}_{t-H:t-1}^\top]^\top\) the history of the previous \(H\) actions and states. Diffusion Policy has been proposed in \cite{chi2023diffusion} to model the multimodal action distribution in robot imitation learning using Denoising Diffusion Probabilistic Models (DDPMs) \cite{ho2020denoising} and Denoising Diffusion Implicit Models (DDIMs) \cite{song2020denoising}. Diffusion Policy regresses a noise prediction function $\bm{\varepsilon}_{\theta}(\mathbf{h}^H_t, \mathbf{a}_t + \bm{\varepsilon}^k, k) = \bm{\varepsilon}^k$ with a network $\bm{\varepsilon}_{\theta}$ parameterized by $\theta$. During training, the current history and actions $(\mathbf{h}_t^H, \mathbf{a}_t)$ are sampled from the demonstrated dataset. Random noise $\bm{\varepsilon}^k$, conditioned on a randomly sampled denoising step $k$, is added to $\mathbf{a}_t$. Thus, the loss can be described as 
\[
\mathcal{L} = \left\| \bm{\varepsilon}_{\theta}(\mathbf{h}^H_t, \mathbf{a}_t + \bm{\varepsilon}^k, k) - \bm{\varepsilon}^k \right\|^2.
\]

During inference, given the history $\mathbf{h}^H_t$, Diffusion Policy executes a sequence of $K$ denoising steps starting from random samples actions $\mathbf{a}_t^k \sim \mathcal{N}(0,1)$ to generate target robot actions $\mathbf{a}_t^0$. This inverse process can be defined as
\[
\mathbf{a}^{k-1} = \alpha \left( \mathbf{a}_t^k - \gamma \bm{\varepsilon}_{\theta}(\mathbf{h}^H_t, \mathbf{a}_t^k, k) + \bm{\varepsilon} \right),
\]
where $\alpha, \gamma$ are the parameters of the noise schedule, $\bm{\varepsilon} \sim \mathcal{N}(0, \sigma^2 I)$. The action $\mathbf{a}_t^0$ can be sampled from the demonstration data or expert policy $\pi: \mathbf{h}^H_t \mapsto \mathbf{a}_t$.


\section{Methodology}\label{sec:meth}
    
\subsection{Problem Definition}

Given a dataset of \(M\) successful demonstrations \(\mathcal{D} = \{ (\mathbf{a}_t, \mathbf{x}_t, \mathbf{h}_t^H)_i \}_{i=1}^{M}\), our objective is to learn a diffusion policy that models the conditional distribution \(p_\pi^{\mathcal{D}}(\mathbf{a}_t \mid \mathbf{x}_t, \mathbf{h}_t^H)\) so that an action is generated as \(\mathbf{a}_t \sim p_\pi^{\mathcal{D}}(\mathbf{a}_t \mid \mathbf{x}_t, \mathbf{h}_t^H)\) using the DDPM approach described in Sec.~\ref{sec:pre_dp} (we omit the zero superscript for clarity). 

However, this formulation requires a very long history (\(H \gg 0\)) to capture all previous failures. To address this limitation, we extend the formulation by conditioning on a set of failure features as
\begin{equation} \label{eq:key_failures}
\mathbf{a}_t \sim p_{\pi}(\mathbf{a}_t \mid \mathbf{x}_t, \mathbf{h}_t^H, \mathbf{z}_{1:N}^f),
\end{equation}
where \(\mathbf{z}^f_i = \mathbf{z}(\mathbf{a}^f_i, \mathbf{x}^f_i)\) extracts key features from the \(i\)-th failure's action \(\mathbf{a}^f_i\) and state \(\mathbf{x}^f_i\), and \(N\) denotes the total number of previous failures. These features may include a learned latent-space representation, key measurements such as forces, or discrete mode indicators. For the purpose of this work, we assume that the function \(\mathbf{z}(\cdot)\) is known, and we discuss different possibilities in Sec.~\ref{sec:meth_z}.

With this formulation, a long history is not required since key features effectively summarize past information. When failures occur, the system extracts these features and conditions future policies on them. However, the resulting input can be high-dimensional, depending on the dimensions of the actions, states, key feature space, and the variable number  of failures (which may range, for example, from 2 to 10). This high dimensionality complicates both model learning and data collection, as it must account for all possible failure combinations.

\subsection{Composition of Diffusion Policies}

To mitigate these challenges and introduce modularity, inspired by \cite{liu2022compositional}, we propose decomposing \eqref{eq:key_failures} into several simpler sub-problems. Hence, \eqref{eq:key_failures} can be rewritten as
\begin{multline}\label{eq:prod_dist}
p_{\pi}(\mathbf{a}_t \mid \mathbf{x}_t, \mathbf{h}_t^H, \mathbf{z}_{1:N}^f) \propto \\ p_a(\mathbf{a}_t)\frac{p_s(\mathbf{a}_t \mid \mathbf{x}_t)}{p_a(\mathbf{a}_t)}\frac{p_h(\mathbf{a}_t \mid \mathbf{h}_t^H)}{p_a(\mathbf{a}t)}\prod_{i=1}^{N}\frac{p_z(\mathbf{a}t \mid \mathbf{z}^f_{i})}{p_a(\mathbf{a}_t)}.
\end{multline}

Using the diffusion model framework, we initialize with a noisy sample and iteratively denoise through the reverse diffusion process
\begin{equation}
    p(\mathbf{a}_t^{k-1} \mid \mathbf{a}_t^{k}) = \mathcal{N}\Big(\alpha\Big(\mathbf{a}_t^k - \gamma\,\hat{\bm{\varepsilon}}(\mathbf{a}_t^k, k)\Big), \sigma_t^2\,\mathbf{I}\Big).
\end{equation}

According to \eqref{eq:prod_dist}, we decompose the denoising term \(\hat{\bm{\varepsilon}}(\mathbf{a}_t^k, k)\) as follows:
\begin{multline}\label{eq:decomposed_diffusion}
 \hat{\bm{\varepsilon}}(\mathbf{a}_t^k, k) = \bm{\varepsilon}_a(\mathbf{a}_t, k) + w_s \Big(\varepsilon_s(\mathbf{a}_t, \mathbf{x}_t, k) - \bm{\varepsilon}_a(\mathbf{a}_t, k)\Big)\\
 + w_h \Big(\bm{\varepsilon}_h(\mathbf{a}_t, \mathbf{h}_t^H, k) - \bm{\varepsilon}_a(\mathbf{a}_t, k)\Big)\\
 + \sum_{i=1}^{N} w_z^i \Big(\bm{\varepsilon}_z(\mathbf{a}_t, \mathbf{z}_i^f, k) - \bm{\varepsilon}_a(\mathbf{a}_t, k)\Big).
\end{multline}

Here, \(w_s\), \(w_h\), and \(w_z^i\) are positive coefficients associated with the state, history, and failure key features, respectively. Note that these coefficients are not necessarily bounded by 1, nor do they sum to 1 (see \cite{liu2022compositional} for details). In the special case where \(w_s = 0\) and \(w_z^i = 0\) for all \(i\) while \(w_h = 1\) (with \(\mathbf{x}_t\) incorporated into \(\mathbf{h}_t^H\)), the formulation reduces exactly to the standard diffusion policy. This modular decomposition offers additional flexibility, enabling the system to accommodate a variable number of failure cases. Consequently, regardless of the number of failures, we need to learn only one model—namely, \(p(\mathbf{a}_t \mid \mathbf{z}^f)\)—to capture the failure behavior, thereby making the learning process more tractable.

In the following, we discuss the effects of the different terms in \eqref{eq:decomposed_diffusion} and how to control their corresponding weights in our scenarios.

\begin{itemize}
\item  \(\bm{\varepsilon}_a(\mathbf{a}_t, k)\): This term encourages sampling actions that are similar to those observed in the demonstrations. It does not incorporate information from the current state, history, or previous failures; rather, it guides the sampling process toward the demonstrated actions.

\item \(w_s\Big(\bm{\varepsilon}_s(\mathbf{a}_t, \mathbf{x}_t, k) - \bm{\varepsilon}_a(\mathbf{a}_t, k)\Big)\): This component steers the actions toward those that match the current state of the robot and its environment. It is analogous to a diffusion policy with zero history. We separate this term from the history component to avoid bias from past failures—especially if a previous action caused a failure—while still promoting meaningful actions considering the current state of the robot. Although relying solely on the current state may lead to less smooth behavior, combining it with the history term allows us to balance responsiveness and smoothness. 

\item \(w_h\Big(\bm{\varepsilon}_h(\mathbf{a}_t, \mathbf{h}_t^H, k) - \bm{\varepsilon}_a(\mathbf{a}_t, k)\Big)\): This term promotes temporal continuity by encouraging the system to follow the action history. In scenarios where a failure occurs, reducing \(w_h\) prevents the system from rigidly adhering to a detrimental history; conversely, in failure-free situations, maintaining \(w_h\) at a level comparable to \(w_s\) ensures smooth motion.

\item \(w_z^i\Big(\bm{\varepsilon}_z(\mathbf{a}_t, \mathbf{z}_i^f, k) - \bm{\varepsilon}_a(\mathbf{a}_t, k)\Big)\): This term allows the system to consider distinct failure cases and, based on the extracted failure features, steer away from regions that led to failures. Unlike the other terms that guide the system toward a desired area, this term acts repulsively. By handling each failure separately, we avoid the need to learn a model for every possible combination or ordering of failures, thus reducing the data collection burden. For details on learning this model, please refer to Sec.~\ref{subsec:data_gathering}.

\end{itemize}

\begin{figure*}[t!]
    \subfigure[$p(\mathbf{a}_t)$]{%
    \includegraphics[width=0.18\textwidth]{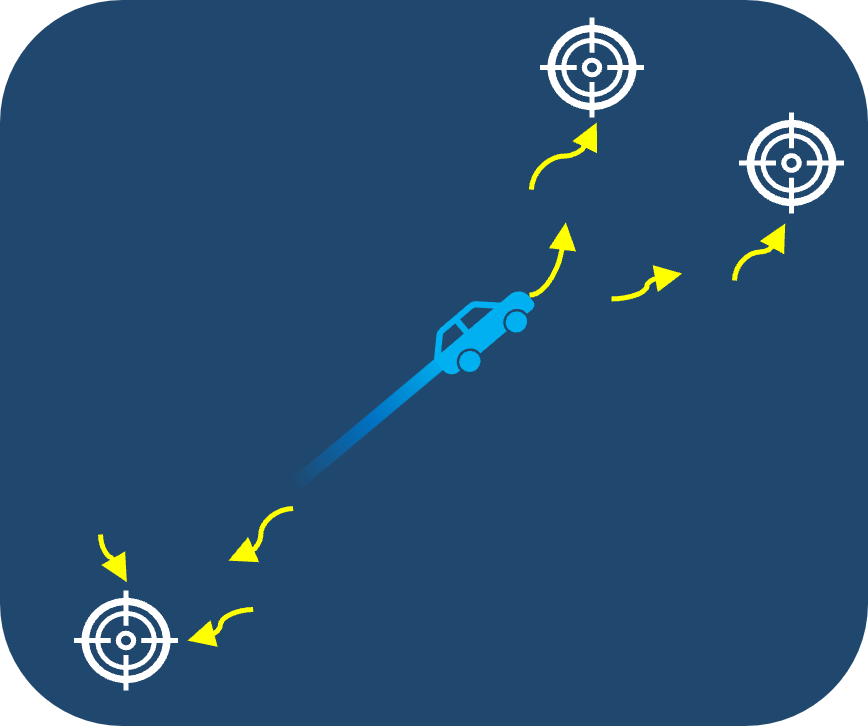}
        }
\subfigure[$p(\mathbf{a}_t|\mathbf{x}_t)$]{    \includegraphics[width=0.18\textwidth]{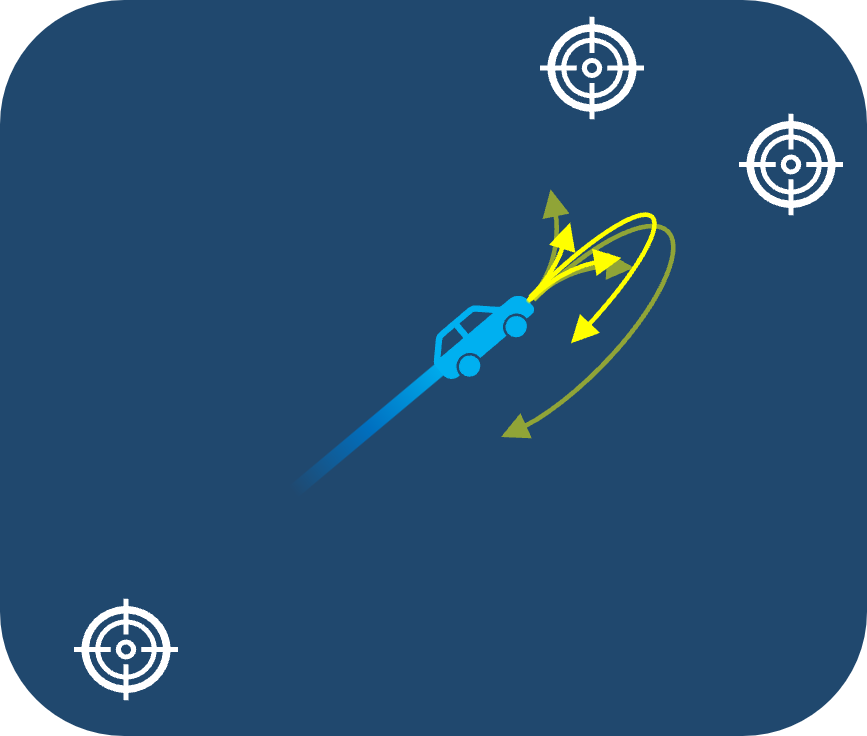}
        } \subfigure[$p(\mathbf{a}_t|\mathbf{h}^H_t)$]{\includegraphics[width=0.18\textwidth]{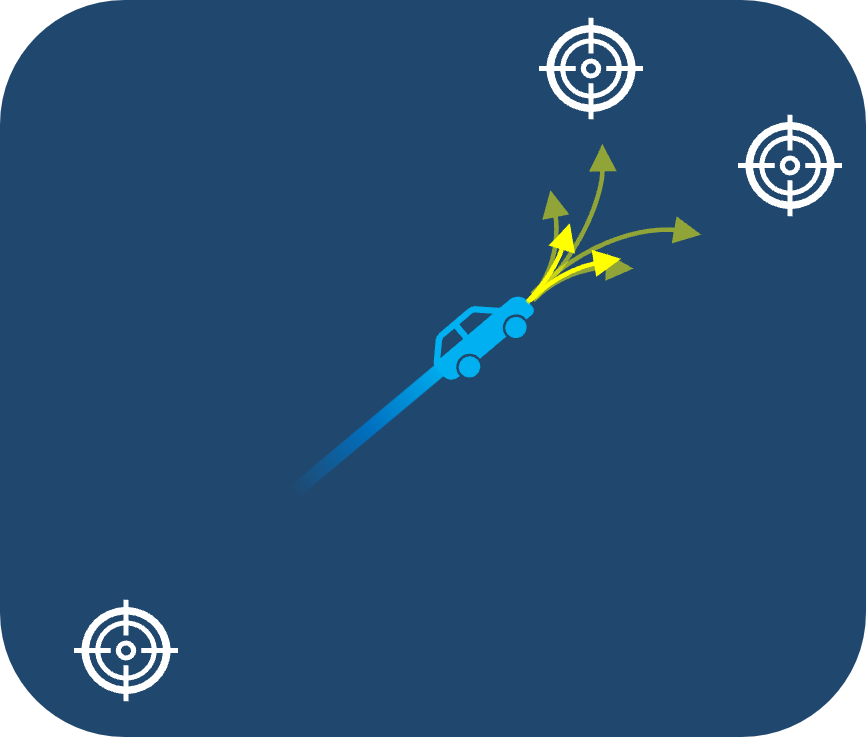}
        }
    \subfigure[$p(\mathbf{a}_t|\mathbf{z}^f_i)$]{%
        \includegraphics[width=0.18\textwidth]{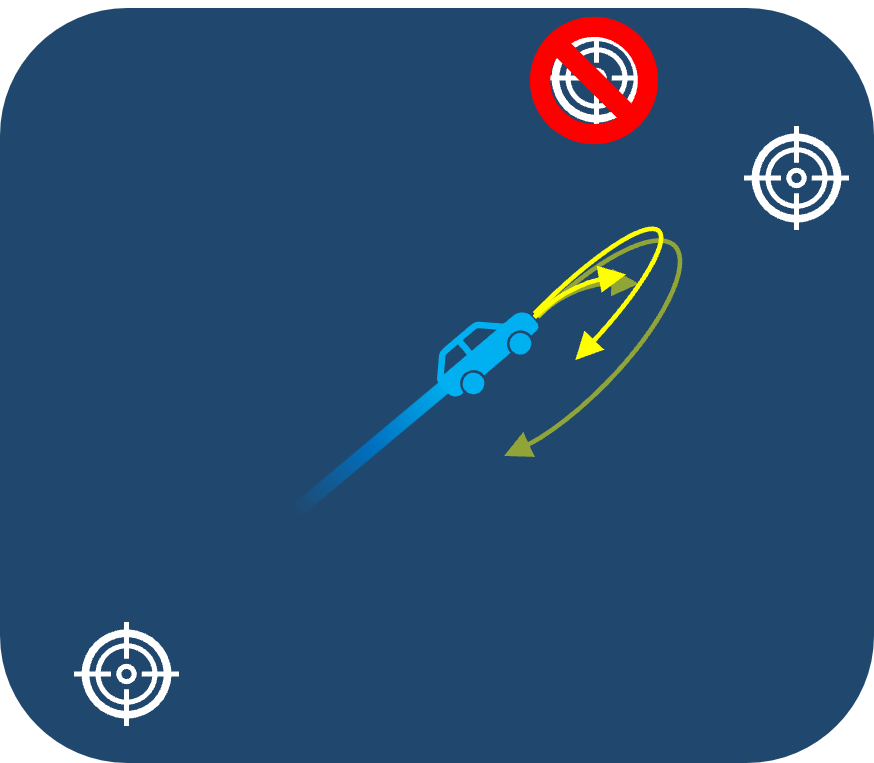}
        }
\subfigure[Combined]{        \includegraphics[width=0.18\textwidth]{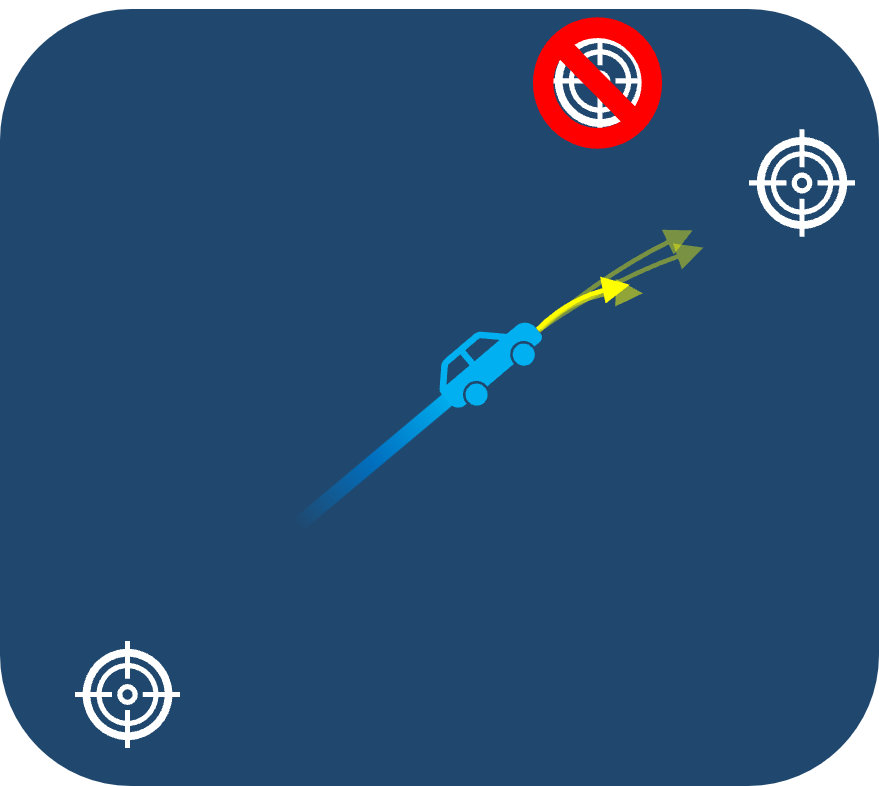}
        }

     \caption{An illustrative example showing samples generated from multiple distributions learned from the same demonstration set. The blue car, with its trajectory history shown as a shaded trail, aims to reach one of the white goal positions, while the yellow arrows indicate sampled actions.}\vspace{-15pt}

     \label{fig:diff_sampling}
\end{figure*}

\subsection{Recovery Model}\label{subsec:data_gathering}
Since we assume that the provided dataset does not include failure cases, we need to synthesize another dataset to train failure recovery. In this regard, we synthesize an auxiliary dataset \(\mathcal{R}\) using distributions \(p(\mathbf{a}_t)\), and \(p(\mathbf{a}_t \mid \mathbf{x}_t)\).

We first relax the recovery definition slightly by considering all actions dissimilar to the failed one as potential recoveries. If we further assume that the cause of failure is static (i.e., repeating the same action in the same state will fail again, as the door hinge location or button location remains unchanged by external factors), then we define the set of recovery actions as
\begin{equation}\label{eq:avoidance_cost}
\mathbf{a} \in \mathcal{R}(\mathbf{z}^f) \quad \text{if} \quad
\begin{cases}
\|\mathbf{z}(\mathbf{a}, \mathbf{x}) - \mathbf{z}(\mathbf{a}^f, \mathbf{x}^f)\|^2 > \delta_z, \\
\|\mathbf{x} - \mathbf{x}^f\|^2 < \delta_x,
\end{cases}
\end{equation}
where \(\mathbf{a}^f\) is the action that led to failure, \(\mathbf{x}^f\) is the state where the failure occurred, \(\delta_z\) is a threshold defining sufficient dissimilarity in the key feature space, and \(\delta_x\) is a threshold defining similarity in the state space. Intuitively, if the state of the system has not changed significantly, we expect the same action to result in failure again; therefore, recovery actions must differ substantially in the failure feature space.

Now we can construct the recovery dataset by traversing the available data and accepting every pair that satisfies \eqref{eq:avoidance_cost}. However, doing so directly on the main dataset can be problematic because the dataset is limited and actions are collected in various states, resulting in a sparse recovery dataset. To address this issue, we perform data synthesis: we randomly select states observed in the demonstrations and, for each state, sample multiple actions such that
\begin{multline}
\mathcal{D}_s(\mathbf{x}_s) = \{ (\mathbf{a}, \mathbf{x}_s) \mid \mathbf{a} \sim \bar{p}^{\mathcal{D}}(\mathbf{a} \mid \mathbf{x}),\; \mathbf{x} \in \mathbf{x}_s + \bm{\xi}_x,\; \\ \bm{\xi}_x \sim \mathcal{N}(\mathbf{0}, \sigma^2 \mathbf{I}) \}.
\end{multline}
Here, \(\bm{\xi}_x\) denotes random noise added to the state. The corresponding noise estimator is defined as
\begin{equation}\label{eq:synth_denois}
     \bar{\bm{\varepsilon}}\big(\mathbf{a}, \mathbf{x}, k\big) = \bm{\varepsilon}_a(\mathbf{a}, k) + w_s \Big(\bm{\varepsilon}_s(\mathbf{a}, \mathbf{x}, k) - \bm{\varepsilon}_a(\mathbf{a}, k)\Big),
\end{equation}
where the key difference between \eqref{eq:synth_denois} and \eqref{eq:decomposed_diffusion} is the exclusion of history in \eqref{eq:synth_denois}. This omission is crucial for two reasons: (1) including history could bias the system toward past actions, thereby reducing the diversity of recovery solutions (see Fig.~\ref{fig:diff_sampling}), and (2) it would lead to a higher-dimensional problem. Additionally, setting \(w_s \leq 1\) encourages the system to explore a broader range of actions by allowing sampling from the entire action space—even those actions that may not be perfectly suited to the current state—while still guiding the samples toward the current state. We found that this diversity is beneficial for learning robust avoidance behavior. During inference, when using \eqref{eq:decomposed_diffusion}, an appropriate value of \(w_s\) is chosen to ensure that the primary actions applied to the robot remain suitable for the current state.

After synthesizing the dataset \(\mathcal{D}_s(\mathbf{x}_s)\) for states sampled from demonstrations, we extract the pairs that satisfy \eqref{eq:avoidance_cost}. With this recovery dataset in hand, we then apply the DDPM method to learn the corresponding distribution.

\begin{figure*}[t!]
   \centering
   \hfill \\
    \subfigure{%
        \includegraphics[width=0.7\textwidth]{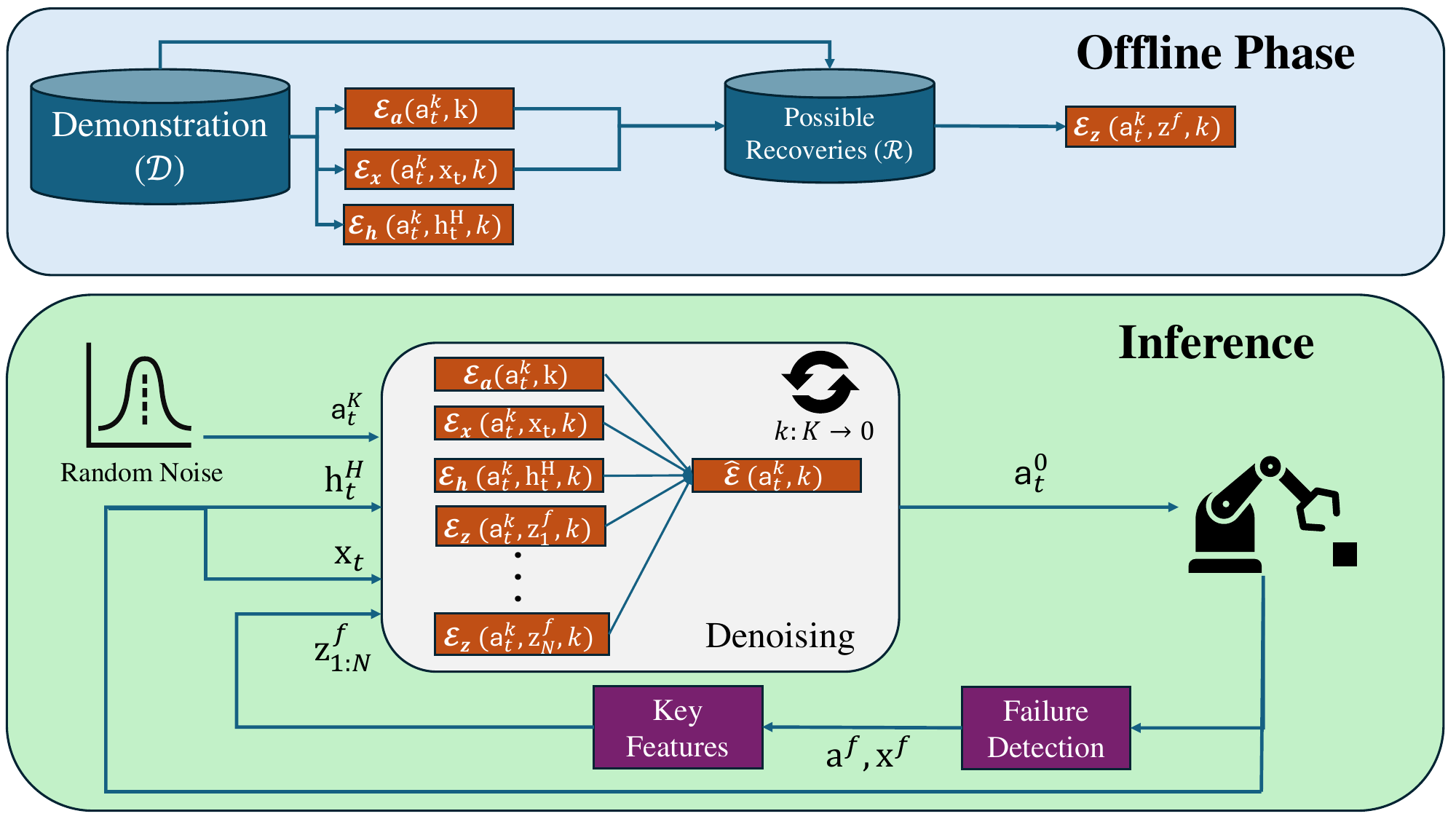}
        }
     \caption{Schematic overview of the proposed method illustrating the offline and inference phases.}

     \label{fig:sum_proces}
\end{figure*} 
\subsubsection{Failure Key Features}\label{sec:meth_z}
The function \(\mathbf{z}(\cdot)\) captures the key features of failed actions. While learning these features autonomously via latent-space methods would be highly beneficial, we leave this for future work. Instead, we discuss three intuitive and practical choices for \(\mathbf{z}(\cdot)\):
\begin{enumerate}
    \item \textbf{Directly using the failed actions:} \(\mathbf{z}(\mathbf{a}^f, \mathbf{x}^f) = \mathbf{a}^f\). In this approach, the robot directly avoids the failed action, which is particularly useful when the state of the system remains largely unchanged after a failure.
    \item \textbf{Using the final state:} \(\mathbf{z}(\mathbf{a}^f, \mathbf{x}^f) = \mathbf{x}^f_T\), where \(\mathbf{x}^f_T\) is the state reached if \(\mathbf{a}^f\) were executed at \(\mathbf{x}^f\); this can be readily extracted from demonstrations.
    \item \textbf{Action primitive:} \(\mathbf{z}(\mathbf{a}^f, \mathbf{x}^f) = m\), where \(m\) is a discrete label indicating the action primitive applied to the system. In this case, the data must be appropriately labeled, or skill discovery methods such as \cite{zhu2022bottom} can be used to obtain these labels.
\end{enumerate}

Fig.~\ref{fig:sum_proces} summarizes the overall process in both the offline and online phases. In the online phase, the robot's behavior is monitored, and when a failure is detected, the corresponding case is added to the failure set. Subsequently, \eqref{eq:decomposed_diffusion} is used to sample a new action guided by previous failures. 

\section{Experiments}\label{sec:exp}
We validated our approach on tasks where some samples failed or underperformed, comparing it with two baselines: a standard diffusion policy (DP) and an enhanced version (DP*), which partitions the solution space and uses rejection sampling to select a new region upon failure.
\begin{figure}[t!]
   \centering
    \subfigure[Object Manipulation]{%
        \includegraphics[width=0.2\textwidth]{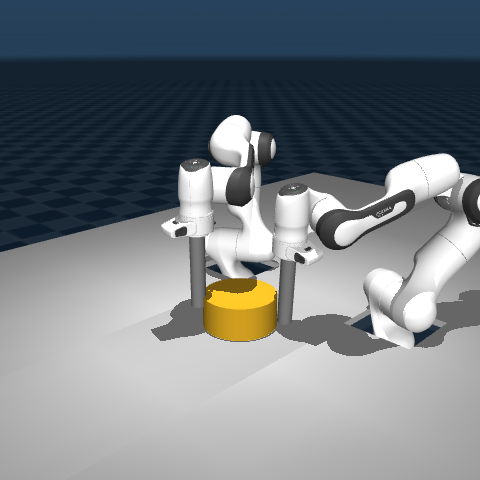}
        }
    \subfigure[Button Pressing]{%
        \includegraphics[width=0.2\textwidth]{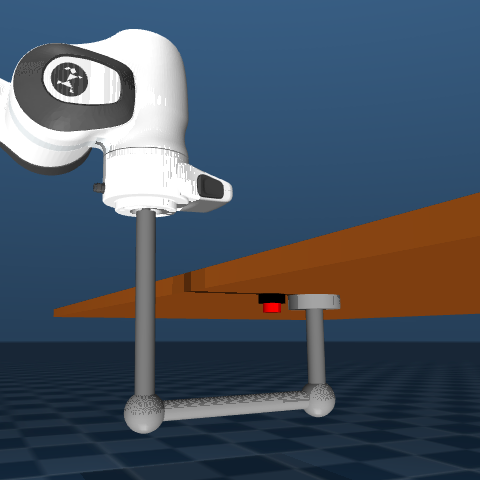}
        }
    \subfigure[Object Packing]{%
        \includegraphics[width=0.28\textwidth]{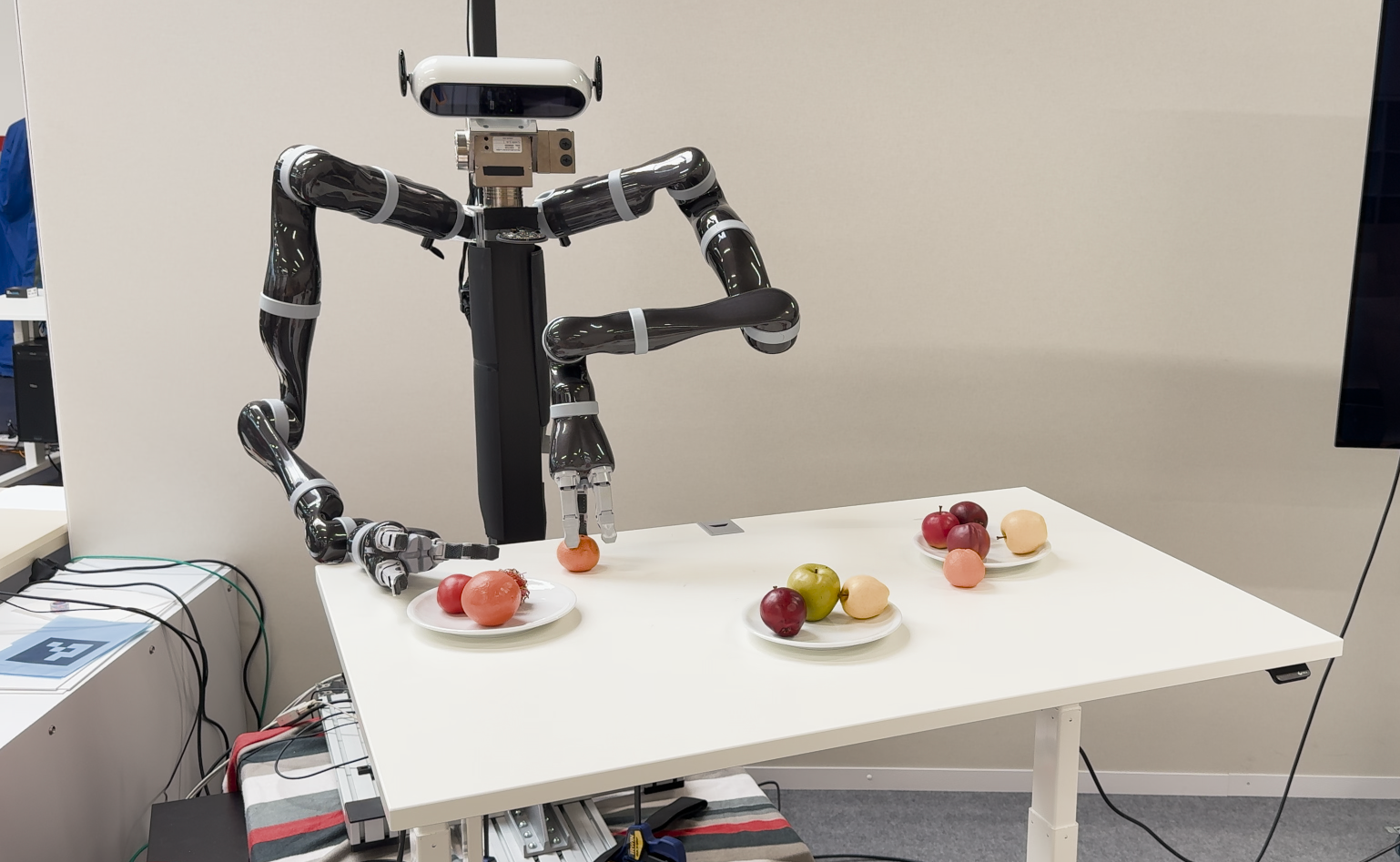}
        }
    \caption{Different experimental setups used in this paper}\vspace{-12pt}
     \label{fig:exp_sets}
\end{figure} 

Our approach does not rely on a high-level planner, so region selection is performed randomly. Additionally, our pipeline requires a failure detection system, which can be implemented using any appropriate method (e.g., \cite{cornelio2024recover}). In our study, demonstrations are collected in a simulated environment using a predefined motion planning method that has access to true parameters, allowing it to perform the task successfully. However, these parameters remain inaccessible to both our approach and the baselines. Importantly, this use of simulation does not violate our assumptions, as the models are not employed during the training phase. 

Performance is evaluated based on both task success rate and the extent to which an implicit objective (e.g., optimality) is achieved. Since one of the aims of imitation learning is to perform tasks without explicitly modeling task or reward functions, some experiments define an implicit objective that is not directly provided to the robot. We then assess how well different methods meet this objective. The following sections detail the experimental setup and results.

In this paper, we adopt the pipeline proposed in \cite{chi2023diffusion}, where at each timestep, the robot observes a history of $H$ past actions and states, predicts a trajectory of length $L$, and executes only the first $p$ steps of that trajectory. This process is repeated until the task is completed. The specific values of $H$, $L$, and $p$ for each experiment are provided in TABLE~\ref{tab:hyper_params}.
\begin{figure*}[t!]
   \centering
    \subfigure[Success Rate]{%
        \includegraphics[width=0.4\textwidth]{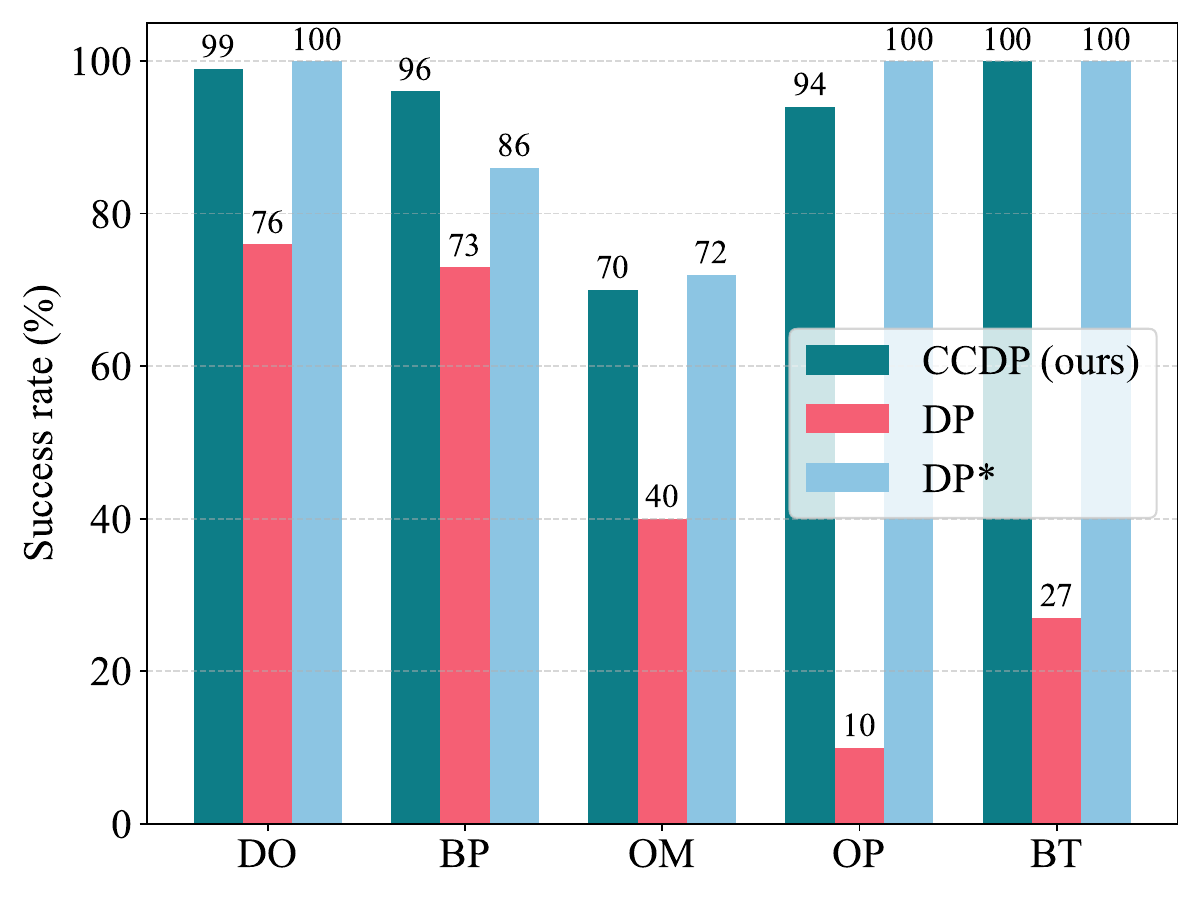}
        }
    \subfigure[Implicit Objective]{%
        \includegraphics[width=0.4\textwidth]{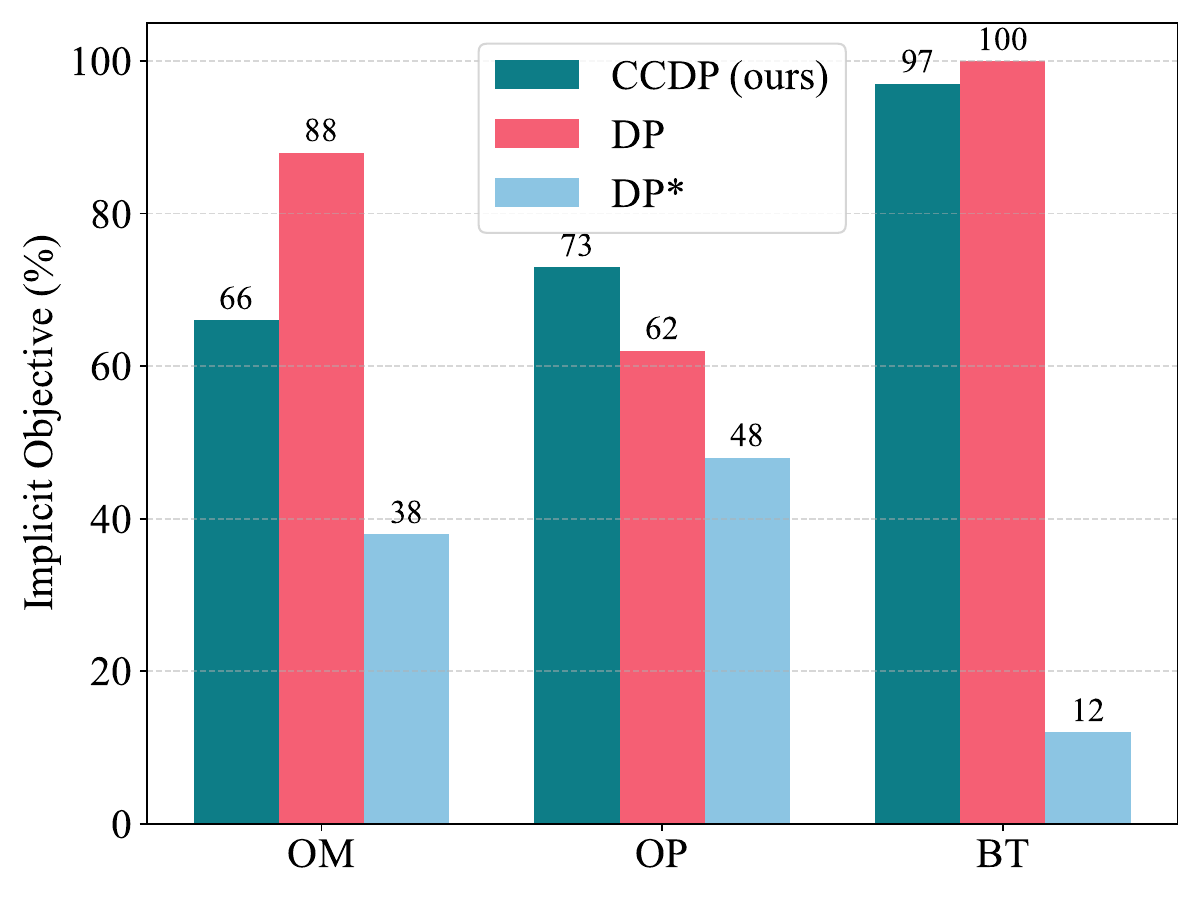}
        }
    \caption{Results from 100 random test scenarios comparing (a) success rates across different tasks and (b) implicit objective fulfillment across various experiments.}\vspace{-15pt}
     \label{fig:res}
\end{figure*}

\begin{table}
    \centering
    \caption{Hyperparameter configurations used in the experiments.}
    \begin{tabular}{c|c|c|c|c|c}
         & BP & DO & OP & OM & BT\\
         \hline
    $\delta_z$ & 0.3 & 0.3 & 0.3 &0.5&0.3\\
    \# history ($H$) & 0 & 2 & 0 & 2 & 2\\
    \# pred ($L$) & 1 & 8 & 8 & 8& 8\\
    \# applied ($p$)& 1 & 8 & 8 & 4 &4\\
    Batch size & 1024 & 32 & 64 & 64 &64\\
    \hline
    \#  hidden layer  &
    [256,256,256]  &
    \multicolumn{4}{c}{[256,512,1024]}\\ 
    \hline
    \# epoches & \multicolumn{5}{c}{100}\\
    noise schedular &
    \multicolumn{5}{c}{Cosine}\\
    denoising step & \multicolumn{5}{c}{100}\\

    \end{tabular}\vspace{-10pt}
    
    \label{tab:hyper_params}
\end{table}

\subsection{Door Opening (DO)}
We consider a door-opening task where the robot must determine the unknown opening direction (e.g., upward, sliding, or pulling) without explicit instructions or prior experiments. In this setting, demonstrations are unlabeled (except for the DP\(^{*}\) baseline), so the robot relies solely on the actions as key features. As shown in Fig.~\ref{fig:res}-a, while the standard DP policy achieves a lower success rate, our method attains performance comparable to DP\(^{*}\) without requiring labeling or explicit access to the door opening direction. An interaction is deemed successful if the robot opens the door within five trials.

\subsection{Button Pressing (BP)}
In this task, the robot must press a button within a predefined area without knowing its exact location. It samples various target points (used as key features), and the task is deemed unsuccessful if the button is not found within 30 attempts. Our method outperforms the baselines without requiring additional discretization. In contrast, the DP\(^{*}\) baseline discretizes the target area into bins, sampling from these regions and removing a bin after a failure; this approach improves over DP but is less efficient than CCDP.

\subsection{Object Manipulation (OM)}
In this task, the robot must move an object from one point to another without knowing its mass. As an implicit objective, the optimal manipulation strategy varies with object weight: light objects are best handled with single-arm pick-and-place, moderately heavy objects require bimanual manipulation, and very heavy objects are pushed. The demonstrations reflect these strategies by using single-hand pick-and-place 70\% of the time, bimanual pick-and-place 20\% of the time, and pushing 10\% of the time. Although the demonstrations do not explicitly specify when each modality should be applied.

During testing, the robot is randomly assigned to one of the mass categories and must select the appropriate manipulation modality in the correct order. Here, the type of primitive serves as the key feature for failure recovery. A standard DP policy, being biased toward the dominant single-hand primitive, yields lower success rates. In contrast, both our method (CCDP) and the DP\(^{*}\) baseline achieve higher success rates, with DP\(^{*}\) slightly outperforming our method. When evaluating how frequently each method selects the appropriate action based on the implicit objective, our method adheres to the predefined order 66\% of the time, compared to 88\% for DP and 38\% for DP$^*$. Here, ''respecting the order'' means that if an object can be picked up with one hand (i.e., it is light), it should not be pushed or picked up bimanually.

\subsection{Objects Packing (OP)}
This task demonstrates that the proposed method extends beyond failure recovery. In Object Packing, the robot must place objects into designated baskets; however, as baskets fill up, it must choose the next closest available one. The demonstrations implicitly encode basket preferences (the closer, the more preferred) through the frequency of occurrence, without an explicit cost function. Here, key avoidance features are derived from the final states of object trajectories. During testing, 12 objects are randomly placed on a table and must be assigned to baskets with a capacity of 4 each. The task is successful if each basket contains exactly 4 objects, and the implicit objective is achieved if the robot consistently selects the nearest available basket.

In this task, the standard DP policy performs poorly because its reliance on random sampling and history conditioning leads to prolonged attempts to use already-filled baskets. In contrast, both the DP\(^{*}\) baseline and our method achieve over 90\% success, with our method notably outperforming DP\(^{*}\) in selecting the closest available basket.

\subsection{Bartender (BT)}
In this final task, we again demonstrate the method’s applicability beyond failure recovery. The robot is taught to fill multiple cups, with demonstrations implicitly prioritizing the nearest cup as the preferred option, followed by the second nearest, and so on. During testing, the robot must fill all cups while we record how often it selects the closest unfilled cup. Once a cup is filled, the sampler is guided to avoid targeting that cup again. The task is considered successful if all cups are filled, and for the implicit objective, we check if the closest available cup at each step is selected.

\section{Discussion}\label{sec:dis}
Our results show that our approach significantly boosts task success compared to DP while preserving the implicit objectives from the demonstration data (unlike DP$^*$). The performance gap between DP and CCDP was expected since DP does not consider past failures—a limitation our method explicitly addresses.

In this discussion, we elucidate the factors contributing to CCDP’s superior performance over DP$^*$ and explore potential avenues for further improvement. Additionally, we identify the current limitations of the approach and propose directions for future work to refine the method.

\subsection{CCDP vs DP$^*$}
Apart from the major difference that DP$^*$ requires the system to be classified before application, whereas CCDP can be directly applied without any classification, another key distinction lies in the nature of the guidance provided by each method. Specifically, DP$^*$ employs a \emph{positive forcing} approach, where the policy is constrained to sample from a specific region. This restriction can limit the system's ability to fully leverage additional sensory and state information. In contrast, CCDP utilizes a \emph{negative forcing} strategy, which restricts the policy from sampling in only one particular area, thereby allowing exploration across multiple regions. This flexibility enables the system to integrate broader contextual cues and converge toward areas that are guided by the available sensory and state information.

\subsection{NOT Operation}
Research such as \cite{liu2022compositional} has demonstrated that, rather than combining multiple distributions via a product as in \eqref{eq:prod_dist})—an operation analogous to a logical AND—it is possible to adopt an alternative approach based on a NOT operation. Hence, instead of learning from a dataset consisting of points that are well separated, the focus shifts to learning from a dataset of points that are closely clustered, after which the NOT operation is applied. This inversion could potentially simplify dataset creation and streamline the learning process.

However, both our experimental experience and the observations reported in \cite{liu2022compositional} indicate that the NOT operation is inherently unstable. Our attempts to implement this strategy did not yield robust results. Consequently, we defer further exploration of this approach to future work. Nonetheless, the proposed method, which employs the composition of diffusion models to refine samples interactively, remains a viable and effective strategy.

\subsection{Limitations and Future Work}
There are two main limitations in the current version of the proposed model. First, failure key features have been defined manually. It would be advantageous to develop an autonomous approach for extracting these features, potentially by leveraging existing work in latent space extraction methods. The second limitation is that we have not yet investigated how to optimally adjust the combination weights in \eqref{eq:decomposed_diffusion} to achieve improved performance. Our experience suggests that although larger weights may enhance exploration, they can also induce system instability. 

As a further avenue for future work, we propose incorporating an offline exploratory mechanism to extract more effective recovery data and policies. This aspect was not considered in the current study, as our primary goal was to develop a purely data-driven approach without reliance on a simulation or a model.

\section{Conclusion}\label{sec:con}
In this paper, we introduced CCDP, a method that provides enhanced control over the action sampling process. Specifically, we demonstrated how the DP method can be decomposed into multiple subproblems, including policies that encourage the system to explore alternative actions beyond those previously executed. This decomposition results in a low-level controller capable of handling failures and guiding sampling process without the need for labeled data or access to explicit failure recovery signals.

Our experimental results indicate that leveraging demonstrations more efficiently through CCDP leads to improved system performance over baseline methods. We validated our approach on a variety of tasks, including object manipulation, packaging, and door opening. We found that it not only increases the success rate but also better adheres to the implicit objectives defined by the demonstrations.

For future work, we plan to investigate methods to optimize the combination of subproblems, automate the extraction of failure key features, and incorporate offline exploratory behaviors to further enhance the quality of recovery options.

\addtolength{\textheight}{-12cm}   




\IEEEtriggeratref{36}
\bibliographystyle{IEEEtran}
\bibliography{bib}



\end{document}